\pdfoutput=1
\documentclass[11pt]{article}

\usepackage[final]{acl}

\usepackage{times}
\usepackage{latexsym}
\usepackage[T1]{fontenc}
\usepackage[utf8]{inputenc}
\usepackage{microtype}
\usepackage{inconsolata}
\usepackage{graphicx}
\usepackage{amsmath,amssymb}
\usepackage{booktabs}
\usepackage{multirow}
\usepackage{url}
\usepackage{float}
\usepackage{adjustbox}

\newcommand{\model}[1]{\textit{#1}}

\title{Anticipating Errors Before Generation: Probing Evidence of Misconception Representations in Frozen LLMs}

\author{
 \textbf{Vivek Hruday Kavuri\thanks{These authors have contributed equally to this work}},
 \textbf{Abraham Paul Elenjical\footnotemark[1]},
 \textbf{Vasudeva Varma}
\\
 IIIT Hyderabad
\\
 \small{
   \textbf{Correspondence:} \href{mailto:kavuri.hruday@research.iiit.ac.in}{\{kavuri.hruday, abraham.elenjical\}@research.iiit.ac.in}
 }
}

\begin{document}
\maketitle

\begin{abstract}
We probe the residual hidden states of six frozen autoregressive LLMs (2B to 9B parameters) for information about student math misconceptions. On 23 misconception categories from the Eedi benchmark, linear probes reach 40.96\%--46.52\% multiclass accuracy (macro-$F_1$ 0.330--0.389), against a uniform-chance macro-$F_1$ of 0.043 and a majority-class accuracy of 9.6\%. Probing the question stem alone, before any answer option is shown, gives comparable decodability (39.66\%--44.63\% accuracy), which is consistent with models encoding likely errors while reading the stem. Across layers, misconception information becomes decodable earlier (relative depth 0.28--0.53) than binary option correctness (0.57--0.96), an ordering that holds in all six models for emergence thresholds of 60--80\% of peak. Non-linear MLP probes do not outperform linear ones, suggesting this information is linearly accessible. Applying the same pipeline to question-level concept labels from ConceptKT/MathQA, we find concepts decodable at macro-$F_1$ 0.729--0.763 (chance 0.04), more so from the stem alone, and once again observed earlier in depth than option correctness. Our evidence is representational: probing shows that the information is present, not that models use it. Code and data splits are available at \url{https://github.com/vivekhruday05/misconception-probes}.

\end{abstract}

\begin{table*}[t]
\centering
\small
\begin{adjustbox}{max width=\textwidth}
\begin{tabular}{llcccccc}
\toprule
\textbf{Model} & \textbf{Task / Condition} & \textbf{Classes} & \textbf{Chance $F_1$} & \textbf{Peak Layer} & \textbf{Peak $F_1$} & \textbf{Peak Acc (\%)} & \textbf{Emergence (80\%)} \\
\midrule
\multirow{4}{*}{\model{Gemma-2-2B}} & Correctness (Stem+Option) & Binary & 0.5000 & 25 & 0.5750 & 74.70\% & Layer 25 (0.96) \\
 & Correct Letter (Stem-Only Control) & 4-Class & 0.2500 & 16 & 0.2619 & 26.49\% & Layer 15 (0.58) \\
 & Misconception (Stem+Option) & 23 & 0.0435 & 26 & \textbf{0.3713} & \textbf{43.51\%} & Layer 11 (0.42) \\
 & Misconception (Stem-Only) & 23 & 0.0435 & 26 & \textbf{0.3515} & \textbf{40.32\%} & Layer 10 (0.38) \\
\midrule
\multirow{4}{*}{\model{Llama-3.2-3B}} & Correctness (Stem+Option) & Binary & 0.5000 & 25 & 0.5881 & 74.29\% & Layer 24 (0.86) \\
 & Correct Letter (Stem-Only Control) & 4-Class & 0.2500 & 10 & 0.2601 & 26.39\% & Layer 10 (0.36) \\
 & Misconception (Stem+Option) & 23 & 0.0435 & 27 & \textbf{0.3492} & \textbf{42.73\%} & Layer 9 (0.32) \\
 & Misconception (Stem-Only) & 23 & 0.0435 & 28 & \textbf{0.3482} & \textbf{44.26\%} & Layer 15 (0.54) \\
\midrule
\multirow{4}{*}{\model{Qwen2.5-3B}} & Correctness (Stem+Option) & Binary & 0.5000 & 28 & 0.6450 & 77.30\% & Layer 25 (0.69) \\
 & Correct Letter (Stem-Only Control) & 4-Class & 0.2500 & 29 & 0.2739 & 27.53\% & Layer 29 (0.81) \\
 & Misconception (Stem+Option) & 23 & 0.0435 & 35 & \textbf{0.3605} & \textbf{45.35\%} & Layer 19 (0.53) \\
 & Misconception (Stem-Only) & 23 & 0.0435 & 35 & \textbf{0.3439} & \textbf{42.08\%} & Layer 22 (0.61) \\
\midrule
\multirow{4}{*}{\model{Qwen2.5-7B}$^\dagger$} & Correctness (Stem+Option) & Binary & 0.5000 & 28 & 0.6952 & 78.02\% & Layer 18 (0.64) \\
 & Correct Letter (Stem-Only Control) & 4-Class & 0.2500 & 25 & 0.2693 & 27.21\% & Layer 19 (0.68) \\
 & Misconception (Stem+Option) & 23 & 0.0435 & 27 & \textbf{0.3301} & \textbf{40.96\%} & Layer 11 (0.39) \\
 & Misconception (Stem-Only) & 23 & 0.0435 & 22 & \textbf{0.3268} & \textbf{39.66\%} & Layer 10 (0.36) \\
\midrule
\multirow{4}{*}{\model{Llama-3.1-8B}$^\dagger$} & Correctness (Stem+Option) & Binary & 0.5000 & 31 & 0.6404 & 74.31\% & Layer 27 (0.84) \\
 & Correct Letter (Stem-Only Control) & 4-Class & 0.2500 & 16 & 0.2647 & 26.56\% & Layer 16 (0.50) \\
 & Misconception (Stem+Option) & 23 & 0.0435 & 32 & \textbf{0.3417} & \textbf{41.62\%} & Layer 9 (0.28) \\
 & Misconception (Stem-Only) & 23 & 0.0435 & 32 & \textbf{0.3081} & \textbf{40.54\%} & Layer 10 (0.31) \\
\midrule
\multirow{4}{*}{\model{Gemma-2-9B}$^\dagger$} & Correctness (Stem+Option) & Binary & 0.5000 & 42 & \textbf{0.7063} & \textbf{78.88\%} & Layer 24 (0.57) \\
 & Correct Letter (Stem-Only Control) & 4-Class & 0.2500 & 27 & 0.2898 & 29.12\% & Layer 24 (0.57) \\
 & Misconception (Stem+Option) & 23 & 0.0435 & 42 & \textbf{0.3887} & \textbf{46.52\%} & Layer 15 (0.36) \\
 & Misconception (Stem-Only) & 23 & 0.0435 & 41 & \textbf{0.3463} & \textbf{44.63\%} & Layer 15 (0.36) \\
\bottomrule
\end{tabular}
\end{adjustbox}
\caption{\textbf{Linear-probe results on Eedi across six LLMs.} Peak macro-$F_1$, peak accuracy, peak layer, and emergence layer (first layer reaching 80\% of peak $F_1$ above chance; relative depth $l/L$ in parentheses). Chance $F_1$ is that of a stratified random predictor. Majority-class accuracy is 75.0\% for binary correctness (25\% of options are correct) and 9.6\% for the 23-way misconception task, so accuracy should be read against these values. $^\dagger$4-bit (NF4) weights.}
\label{tab:results}
\end{table*}

\section{Introduction}
Large Language Models (LLMs) are increasingly used as automated tutors and reasoning assistants \cite{eedi-mining-misconceptions-in-mathematics,https://doi.org/10.1111/ejed.12533, 10.1145/3569576}. A useful tutor must recognise not only \emph{that} a student is wrong but \emph{how}: the specific misconception behind an error. This raises an interpretability question: when an LLM reads a math problem, do its hidden states already carry information about the errors a student is likely to make, or does this information surface only once an erroneous answer is shown?

As a concrete example, consider the Eedi question \emph{``Solve $2n+3 \le 3n-7$''}. The distractor $2 \le n$ is annotated with the misconception \emph{``When solving an equation, uses the same operation rather than the inverse''}, the most frequent category in our data. We ask whether a linear probe on the model's hidden state can recover this label, both after reading stem and distractor, and after reading only the stem.

\begin{figure}[t]
  \centering
  \includegraphics[width=\linewidth]{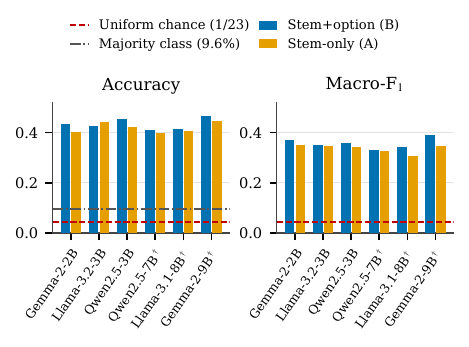}
  \caption{\textbf{Misconception decodability on Eedi (linear probes, peak layer).} Accuracy (left) is compared against the majority-class baseline (9.6\%) and uniform chance (4.35\%); macro-$F_1$ (right) against uniform chance (0.0435). $^\dagger$4-bit weights.}
  \label{fig:cross_model}
\end{figure}

We conduct a layer-wise probing study of six open LLMs (\model{Gemma-2-2B}, \model{Llama-3.2-3B}, \model{Qwen2.5-3B}, \model{Qwen2.5-7B}, \model{Llama-3.1-8B}, \model{Gemma-2-9B}) \cite{gemmateam2024gemma2improvingopen, grattafiori2024llama3herdmodels, qwen2025qwen25technicalreport} on the Eedi benchmark \cite{eedi-mining-misconceptions-in-mathematics}, and apply the same pipeline to question-level concept labels from ConceptKT \cite{kang-etal-2026-conceptkt} on MathQA word problems \cite{amini-etal-2019-mathqa}. We ask four research questions:
\begin{itemize}
    \setlength{\itemsep}{0pt}\setlength{\parskip}{0pt}\setlength{\parsep}{0pt}
    \item \textbf{RQ1 (Existence):} Is the misconception behind a wrong answer linearly decodable from frozen hidden states?
    \item \textbf{RQ2 (Specificity):} Is the decoded information fine-grained enough to separate many misconception categories?
    \item \textbf{RQ3 (Dissociation):} Do misconception and correctness information become decodable at different depths?
    \item \textbf{RQ4 (Anticipation vs.\ Recognition):} Is misconception information decodable from the question stem alone, before the answer options are revealed?
\end{itemize}

Our contributions are: (1) misconception categories are linearly decodable well above chance and majority baselines in all six models; (2) stem-only probes reach decodability comparable to stem+option probes; (3) misconception information becomes decodable at shallower depth than option correctness, robustly across emergence thresholds of 60--80\%; and (4) the same three patterns hold for question-level concept labels on ConceptKT/MathQA. Throughout, our evidence is \emph{representational}: probing establishes that information is linearly present, not that the model uses it when generating. Causal tests (steering, patching) are left to future work.

\section{Related Work}
\paragraph{Probing LLM Representations.}
Linear classifier probes \cite{alain2018understandingintermediatelayersusing} inspect linguistic and semantic knowledge in transformers \cite{tenney-etal-2019-bert,hewitt-manning-2019-structural,belinkov-2022-probing}. Recent work probes internal representations of truth \cite{burns2024discoveringlatentknowledgelanguage,azaria2023internalstatellmknows,marks2024geometrytruthemergentlinear}. Prior work mostly targets factual truthfulness rather than structured human error categories.

\paragraph{From Probes to Interventions.}
Linear directions found by probing can be used to intervene: Inference-Time Intervention \cite{li2023inferencetime} adds probe-derived directions to attention heads to increase truthfulness, and activation addition \cite{turner2023activationaddition,rimsky-etal-2024-steering} and representation engineering \cite{zou2025representationengineeringtopdownapproach} steer behaviour along contrastive directions. Whether misconception directions support similar interventions is untested; our results identify candidate directions and layers for such work.

\paragraph{Layer-wise Emergence.}
Mechanistic interpretability aims to reverse-engineer transformer computation \cite{elhage2021mathematical}. Studies of factual recall \cite{meng2023locatingeditingfactualassociations,geva-etal-2023-dissecting} find that information is extracted in early-to-middle layers before projection to the vocabulary. We ask whether error categories show a similar early availability.

\paragraph{Student Misconceptions in Educational AI.}
Modeling misconceptions is central to intelligent tutoring \cite{https://doi.org/10.1111/ejed.12533, 10.1145/3569576}. Eedi \cite{eedi-mining-misconceptions-in-mathematics} provides diagnostic math questions with annotated misconceptions, and ConceptKT \cite{kang-etal-2026-conceptkt} annotates MathQA word problems with the concepts they require for concept-level knowledge tracing. Prior work evaluates LLMs as end-to-end predictors; we instead probe their internal activations.

\begin{figure*}[t]
  \centering
  \includegraphics[width=0.92\linewidth]{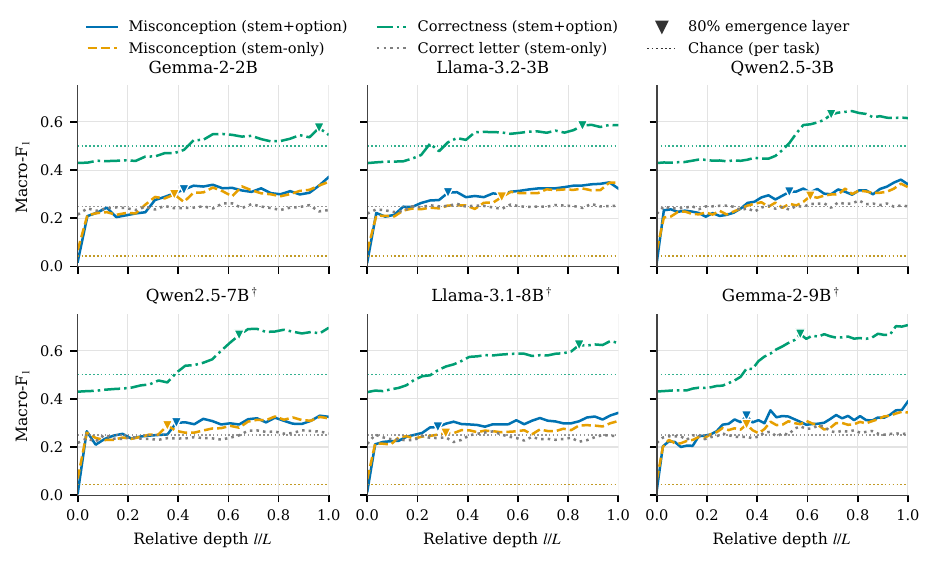}
  \caption{\textbf{Layer-wise linear-probe macro-$F_1$ on Eedi across relative depth $l/L$.} Dotted horizontal lines mark each task's chance level; $\blacktriangledown$ marks the 80\% emergence layer. In every model the misconception marker (blue) precedes the correctness marker (green). Thresholds of 60/70/90\% are reported in Table~\ref{tab:app_thresholds} (Appendix~\ref{sec:app_emergence}). $^\dagger$4-bit weights.}
  \label{fig:trajectories}
\end{figure*}

\section{Methodology}

\subsection{Dataset \& Task Formulations}
Eedi contains 1,869 questions (7,476 answer choices) spanning 757 constructs (topics); each question has a stem, four options (A--D), and misconception annotations on distractors. Of the 4,370 distractors carrying a misconception label, there are 1,604 distinct misconception IDs; exactly 23 have at least 15 instances, giving 565 instances. We chose $\geq 15$ as the minimum that yields roughly three test instances per class per fold under 5-fold cross-validation, so the 23 categories are fully determined by this threshold. We define three tasks:
\begin{enumerate}
    \setlength{\itemsep}{0pt}\setlength{\parskip}{0pt}\setlength{\parsep}{0pt}
    \item \textbf{Misconception} ($N{=}565$, 23 classes): predict the misconception of a distractor. Uniform chance $F_1 \approx 0.0435$; majority-class accuracy 9.6\%.
    \item \textbf{Binary correctness} ($N{=}7{,}476$): predict whether an option is correct. Chance $F_1 = 0.5$; majority-class accuracy 75\%.
    \item \textbf{Correct-letter control} ($N{=}1{,}869$): predict the correct letter from the stem alone (chance 0.25).
\end{enumerate}
\paragraph{ConceptKT/MathQA.} To test the pipeline on a second dataset and question format, we use ConceptKT, whose question IDs index MathQA multi-step word problems with five options. We link 4,036 questions (20,180 choices) and label each question with its primary associated concept, keeping the 25 most frequent concepts (3,846 questions). All four distractors of a question carry its concept label (15,384 instances; uniform chance $F_1$ 0.04, majority-class accuracy 15.8\%); binary correctness uses all choices (majority-class accuracy 80\%), and the correct-letter control is 5-way (chance 0.20). Class lists and a worked example are in Appendix~\ref{sec:app_data}.

\subsection{Prompt Conditions \& Cross-Validation}
We extract residual-stream hidden states $h_l \in \mathbb{R}^d$ at every layer. In \textbf{Condition B} (stem+option, ``recognition'') the prompt contains a topic line, the stem, and one option, and we read the last option token. The topic line holds the construct name on Eedi and the coarse MathQA category (e.g.\ \emph{physics}, \emph{gain}) on ConceptKT, so it never contains the label. In \textbf{Condition A} (stem-only, ``anticipation'') the option is omitted and we read the last stem token; each distractor instance is then represented by its question's stem activation (templates in Appendix~\ref{sec:app_prompts}). All experiments use 5-fold stratified group cross-validation, grouped by construct on Eedi (test questions come from unseen topics) and by question on ConceptKT (all options of a question share a fold).

\subsection{Probes \& Metrics}
The linear probe is $L_2$-regularised logistic regression with inverse regularisation strength $C{=}1.0$; the non-linear probe is a 2-layer MLP (hidden size 128, dropout 0.1). We report macro-$F_1$ (primary) and accuracy at the peak layer and the \emph{emergence layer}: the first layer whose macro-$F_1$ reaches $\text{chance} + r\cdot(\text{peak}-\text{chance})$, with $r{=}0.8$ by default; relative depth is $l/L$ for a model with $L$ blocks. The three largest models are loaded with 4-bit NF4 \cite{NEURIPS2023_1feb8787} weights; their activations are stored in fp16 but are computed through quantized weights (Appendix~\ref{sec:app_hardware}).

\section{Results}

\subsection{Existence \& Specificity (RQ1, RQ2)}
Across all six models (Table~\ref{tab:results}, Figure~\ref{fig:cross_model}), Condition-B linear probes reach 40.96\%--46.52\% accuracy and 0.330--0.389 macro-$F_1$ on 23 classes, i.e.\ 4.3--4.9$\times$ the majority-class accuracy and 7.6--8.9$\times$ uniform-chance macro-$F_1$. Because folds are grouped by construct, probes must generalise to unseen topics. Macro-$F_1$ well above chance shows that the signal is not carried by a few frequent classes and that the representations separate many fine-grained categories (\textbf{RQ1, RQ2}). On ConceptKT (Table~\ref{tab:conceptkt}), question-level concepts are decodable at macro-$F_1$ 0.729--0.763 (81.6\%--84.2\% accuracy), 18--19$\times$ chance. Probing alone cannot rule out that probes partly exploit correlated surface features of the question or option; a logit-lens comparison \cite{nostalgebraist2020logitlens} is a complementary control we leave to future work.

\begin{figure}[t]
  \centering
  \includegraphics[width=\linewidth]{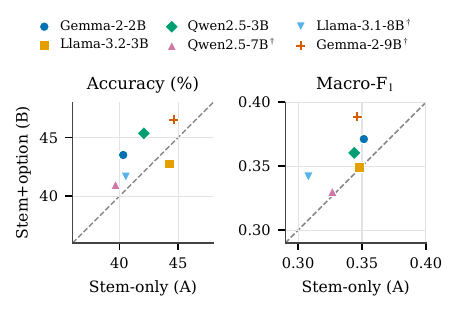}
  \caption{\textbf{Anticipation (stem-only, A) vs.\ recognition (stem+option, B) on Eedi.} Peak linear-probe accuracy and macro-$F_1$ per model; the dashed line is $y{=}x$. $^\dagger$4-bit weights.}
  \label{fig:anticipation}
\end{figure}

\begin{table}[t]
\centering
\small
\setlength{\tabcolsep}{3.5pt}
\begin{adjustbox}{max width=\columnwidth}
\begin{tabular}{l ccc cc}
\toprule
& \multicolumn{3}{c}{\textbf{Concept} (25 classes)} & \multicolumn{2}{c}{\textbf{Correctness}} \\
\cmidrule(lr){2-4}\cmidrule(lr){5-6}
\textbf{Model} & B & A & Em. & $F_1$ & Em. \\
\midrule
\textit{Chance} & \textit{0.040} & \textit{0.040} & & \textit{0.500} & \\
\model{Gemma-2-2B} & 0.734 & 0.786 & 0.23 & 0.477 & n/a \\
\model{Llama-3.2-3B} & 0.729 & 0.781 & 0.14 & 0.521 & 1.00 \\
\model{Qwen2.5-3B} & 0.741 & 0.771 & 0.33 & 0.519 & 0.69 \\
\model{Qwen2.5-7B}$^\dagger$ & 0.763 & 0.789 & 0.18 & 0.638 & 1.00 \\
\model{Llama-3.1-8B}$^\dagger$ & 0.734 & 0.788 & 0.16 & 0.580 & 0.94 \\
\model{Gemma-2-9B}$^\dagger$ & 0.755 & 0.797 & 0.17 & 0.634 & 0.95 \\
\bottomrule
\end{tabular}
\end{adjustbox}
\caption{\textbf{Linear-probe results on ConceptKT/MathQA.} Peak macro-$F_1$ for the 25-way concept label under stem+option (B) and stem-only (A) prompts, and for binary option correctness (Condition B; majority-class accuracy 80\%). Em.: relative depth $l/L$ at which the Condition-B probe first reaches 80\% of its peak above chance; n/a: peak not above chance. $^\dagger$4-bit weights.}
\label{tab:conceptkt}
\end{table}

\subsection{Stem-Only Anticipation (RQ4)}
Probes on the stem alone reach 39.66\%--44.63\% accuracy and 0.308--0.351 macro-$F_1$ (Figure~\ref{fig:anticipation}). Compared with Condition B, stem-only accuracy is lower by 1.1--3.3 points in five models and higher by 1.5 points in \model{Llama-3.2-3B}; macro-$F_1$ is lower by 0.001--0.042. Thus most of the misconception information decodable after seeing an option is already decodable from the stem. This is consistent with models encoding which errors a question is likely to induce while reading it (\textbf{RQ4}). The stem-only correct-letter control stays near chance (0.260--0.290 vs.\ 0.25), so the stem representation does not simply encode the answer key. On ConceptKT, stem-only concept probes are \emph{better} than stem+option probes in all six models (0.771--0.797; $+$0.027 to $+$0.054), consistent with the label being a property of the question rather than of the option; the correct-letter control again stays near chance (0.231--0.236 vs.\ 0.20). Part of the stem-only signal may reflect question-level properties that correlate with misconception type, which is expected when anticipating misconceptions from the stem alone.

\subsection{Dissociation Across Depth (RQ3)}
At the 80\% threshold, Condition-B misconception probes emerge at relative depth 0.28--0.53 (e.g.\ layer 9 of 32 in \model{Llama-3.1-8B}, 15 of 42 in \model{Gemma-2-9B}) and continue to improve into the final layers; binary correctness emerges at 0.57--0.96 (layers 27 and 24, respectively; Figure~\ref{fig:trajectories}). Correctness is also the harder target: its accuracy (74.29\%--78.88\%) is close to the 75\% majority rate, and its macro-$F_1$ (0.575--0.706) is moderately above chance. Within every model, misconception emerges before correctness at thresholds of 60\%, 70\%, and 80\%; at 90\% the ordering holds in four of six models, reversing in \model{Qwen2.5-3B} and \model{Gemma-2-9B}, whose misconception curves keep rising until the last layers (Table~\ref{tab:app_thresholds}). The ordering holds for both fp16 and 4-bit models. On ConceptKT, concept probes emerge at 0.14--0.33 and correctness at 0.69--1.00; correctness accuracy (80.1\%--82.0\%) is close to the 80\% majority rate, and for \model{Gemma-2-2B} correctness never exceeds chance. In the other five models the concept emerges before correctness at every threshold from 60\% to 90\%. These results indicate that misconception-relevant information becomes linearly available earlier in the forward pass than correctness information (\textbf{RQ3}); they do not show where, or whether, the model computes either.

\subsection{Probe Capacity \& Model Scale}
MLP probes do not perform better than linear probes (differences between $-$0.024 and $+$0.012 macro-$F_1$; linear $\ge$ MLP in 11 of 12 model--condition pairs; Appendix~\ref{sec:app_parity}), suggesting that misconception information is linearly accessible. On ConceptKT, the MLP is slightly better in all 12 pairs, but only by 0.007--0.032 (mean 0.022), so concept information is also largely linearly accessible. Peak Condition-B macro-$F_1$ on Eedi is 0.371 (\model{Gemma-2-2B}), 0.349 (\model{Llama-3.2-3B}), 0.361 (\model{Qwen2.5-3B}), 0.330 (\model{Qwen2.5-7B}), 0.342 (\model{Llama-3.1-8B}), and 0.389 (\model{Gemma-2-9B}): the largest model is best, but the trend is not monotonic, with dips at 7B/8B (Appendix~\ref{sec:app_scale}). Because the larger models are also quantized, we do not attribute these differences to scale alone.

\section{Conclusion}
Linear probes on six frozen LLMs recover 23 Eedi student-misconception categories well above chance and majority baselines, almost as well from the question stem alone as from stem plus distractor. Misconception information becomes linearly decodable at shallower depth than option correctness, robustly across emergence thresholds. The same patterns hold for question-level concept labels on ConceptKT/MathQA. These findings are representational: they locate where misconception-relevant information is linearly available, and they identify candidate directions for future causal work on monitoring and steering in educational AI.

\section*{Limitations}
\begin{enumerate}
    \setlength{\itemsep}{0pt}\setlength{\parskip}{0pt}\setlength{\parsep}{0pt}
    \item \textbf{Representation vs.\ use.} Probing establishes that information is linearly decodable, not that the model uses it. Causal methods such as activation steering \cite{li2023inferencetime,rimsky-etal-2024-steering} or patching are needed to test functional relevance.
    \item \textbf{Baselines.} We compare against uniform-chance and majority-class baselines. A logit-lens baseline \cite{nostalgebraist2020logitlens} would test whether surface next-token information alone predicts the labels; we leave this to future work.
    \item \textbf{Label support.} Only 23 of 1,604 Eedi misconception IDs meet our support threshold (565 of 4,370 labelled distractors). Long-tail misconceptions remain untested.
    \item \textbf{Student-level labels.} Our ConceptKT labels are question-level concepts. The public ConceptKT release also records the concepts individual students were missing, but only for 463 responses from six students and without the option each student chose, so it cannot support a distractor-level analysis like Eedi's; linking these responses to the chosen answers is a natural next step.
    \item \textbf{Quantization.} The three largest models use 4-bit weights; we did not run a full-precision ablation, so precision and scale are confounded in cross-model comparisons.
    \item \textbf{Popularity data.} Distractor-popularity probing was not possible because the public Eedi release lacks student response frequencies.
\end{enumerate}

\section*{Ethical Considerations}
\begin{enumerate}
    \setlength{\itemsep}{0pt}\setlength{\parskip}{0pt}\setlength{\parsep}{0pt}
    \item \textbf{Educational Fairness \& Bias:} Automated error modeling tools in education must be evaluated across diverse student groups to prevent systematic bias in error diagnosis.
    \item \textbf{Pedagogical Safety:} Interventions on internal representations in tutoring systems must be validated to avoid incorrect feedback or reinforcing misconceptions.
    \item \textbf{Dual-Use Considerations:} Knowledge of latent error representations could in principle be used to craft misleading distractors; our aim is transparency and better diagnostic tutoring.
    \item \textbf{Artifacts:} We use the public Eedi, ConceptKT, and MathQA releases and open-weight models under their respective licenses; no new personal data is collected.
\end{enumerate}

\bibliography{custom}

\clearpage
\appendix

\section*{Appendix}
\addcontentsline{toc}{section}{Appendix}
This appendix gives the details needed to reproduce and interpret the results in the main paper. \textbf{Appendix~\ref{sec:app_setup}} describes the experimental setup: compute and quantization, the evaluated models, prompt templates, probe objectives, and the cross-validation, baseline, and emergence definitions. \textbf{Appendix~\ref{sec:app_data}} describes how the Eedi and ConceptKT/MathQA data were constructed, lists all label classes, and walks through a worked example. \textbf{Appendix~\ref{sec:app_results}} reports additional results: layer-wise curves with cross-validation variance on Eedi, complete ConceptKT results and curves, emergence under alternative thresholds, the linear vs.\ MLP comparison, and the relation to model size. \textbf{Appendix~\ref{sec:app_ai}} describes the use of AI assistants.

\section{Experimental Setup Details}
\label{sec:app_setup}

\subsection{Compute and Quantization}
\label{sec:app_hardware}
All the experiments (feature extraction, probing, and cross-validation) ran on a single node of an HPC cluster managed by SLURM, with 4$\times$ NVIDIA GeForce RTX 2080 GPUs (10\,GB each), using CUDA 12.1, PyTorch 2.1.2, Hugging Face \texttt{transformers} 4.40.1, \texttt{bitsandbytes} 0.43.0, and scikit-learn 1.3.0; models were processed one at a time, each using all four GPUs.

\model{Qwen2.5-7B}, \model{Llama-3.1-8B}, and \model{Gemma-2-9B} do not fit in fp16 on the RTX 2080 node and were loaded with 4-bit NF4 weights (double quantization, fp16 compute); we kept the same setting for ConceptKT so that both datasets are processed identically. Their hidden states are saved in fp16, like those of the three smaller fp16 models, but they are produced by a forward pass through quantized weights, so quantization can affect the representations themselves. The two precision groups give qualitatively the same picture: the misconception-before-correctness ordering holds in both (Table~\ref{tab:app_thresholds}), and the peak decodability of the 4-bit models lies within the range of the fp16 models (Table~\ref{tab:results}).

\subsection{Evaluated Models}
\label{sec:app_models}
Table~\ref{tab:app_model_specs} lists the six instruction-tuned models, which span three families and 2.6B--9.2B parameters. Relative depths are computed as $l/L$, where $L$ is the number of transformer blocks; layer $0$ is the embedding output, so each model contributes $L{+}1$ probed layers.

\begin{table}[!htbp]
\centering
\small
\begin{adjustbox}{max width=\columnwidth}
\begin{tabular}{lcccccc}
\toprule
\textbf{Model} & \textbf{Params} & \textbf{$L$} & \textbf{$d_{\text{model}}$} & \textbf{Heads} & \textbf{Context} & \textbf{Weights} \\
\midrule
\model{Gemma-2-2B-it} & 2.6B & 26 & 2304 & 8 & 8192 & fp16 \\
\model{Llama-3.2-3B-Instruct} & 3.2B & 28 & 3072 & 24 & 131072 & fp16 \\
\model{Qwen2.5-3B-Instruct} & 3.0B & 36 & 2048 & 16 & 32768 & fp16 \\
\model{Qwen2.5-7B-Instruct} & 7.6B & 28 & 3584 & 28 & 32768 & 4-bit NF4 \\
\model{Llama-3.1-8B-Instruct} & 8.0B & 32 & 4096 & 32 & 131072 & 4-bit NF4 \\
\model{Gemma-2-9B-it} & 9.2B & 42 & 3584 & 16 & 8192 & 4-bit NF4 \\
\bottomrule
\end{tabular}
\end{adjustbox}
\caption{\textbf{Evaluated models.} $L$ = number of transformer blocks.}
\label{tab:app_model_specs}
\end{table}

\subsection{Prompt Templates and Read-out Positions}
\label{sec:app_prompts}
Both conditions use plain-text templates without a chat template.

\paragraph{Condition B (stem+option):}
\begin{quote}
\small\texttt{Construct: \{ConstructName\}\textbackslash n\\Question: \{QuestionText\}\textbackslash n\\Option (\{Letter\}): \{OptionText\}}\normalsize
\end{quote}
The hidden state is read at the final token of the option text, at every layer. One prompt is built per answer choice, so the binary-correctness task uses all answer choices (7,476 on Eedi, 20,180 on ConceptKT) and the label task uses the labelled distractors (565 and 15,384).

\paragraph{Condition A (stem-only):}
\begin{quote}
\small\texttt{Construct: \{ConstructName\}\textbackslash n\\Question: \{QuestionText\}}\normalsize
\end{quote}
The hidden state is read at the last token of the stem. One prompt is built per question; every distractor instance of the misconception task is paired with its question's stem activation. Because no option is present, a Condition-A probe can only use information available before any option is read. When two distractors of the same question carry different misconceptions (53 of the 391 questions in the misconception task), the stem-only probe cannot separate them, which bounds Condition-A accuracy below 100\%. On Eedi the \texttt{Construct} field holds the construct name, which describes the topic of a question (e.g.\ \emph{Solve linear inequalities in one variable}) and never names a misconception. On ConceptKT it holds the MathQA problem category (one of \emph{gain}, \emph{general}, \emph{geometry}, \emph{other}, \emph{physics}, \emph{probability}), which is much coarser than the 25 concept labels; the concept name itself never appears in the prompt template.

\subsection{Probe Objectives}
\label{sec:app_formulation}
\paragraph{Linear probe.}
For $K$ classes and input $h \in \mathbb{R}^{d}$, logistic regression with parameters $W \in \mathbb{R}^{K \times d}$, $b \in \mathbb{R}^K$ minimises
\begin{equation}
\begin{split}
\mathcal{L}(W, b) = &-\frac{1}{N} \sum_{i=1}^N \sum_{k=1}^K y_{i,k} \log \frac{e^{W_k^\top h_i + b_k}}{\sum_{j=1}^K e^{W_j^\top h_i + b_j}} \\
&+ \frac{1}{2C} \|W\|_F^2
\end{split}
\end{equation}
where $C{=}1.0$ is the inverse regularisation strength (smaller $C$ means stronger regularisation), solved with L-BFGS (scikit-learn, at most 200 iterations). Features are standardised using training-fold statistics.

\paragraph{MLP probe.}
\begin{equation}
\begin{split}
h^{(1)}_i &= \text{ReLU}(W_1 h_i + b_1), \\
\hat{y}_i &= \text{softmax}(W_2 \text{Dropout}(h^{(1)}_i) + b_2)
\end{split}
\end{equation}
with $W_1 \in \mathbb{R}^{128 \times d}$, $W_2 \in \mathbb{R}^{K \times 128}$, and dropout $p{=}0.1$, trained with AdamW (lr $10^{-3}$, weight decay $10^{-4}$, batch size 64) for a fixed 20 epochs, without early stopping. For binary correctness the output layer has a single logit trained with a sigmoid (binary cross-entropy) loss. A separate probe is trained for every layer, task, condition, and fold.

\subsection{Cross-Validation, Baselines, and Emergence}
\label{sec:app_cv}
\paragraph{Grouped cross-validation.}
Let $\mathcal{G} = \{G_1, \dots, G_M\}$ be the groups: the $M{=}757$ Eedi constructs, or the ConceptKT questions. Folds $F_1,\dots,F_5$ partition the data, every instance of a group belongs to exactly one fold, and class proportions are stratified as far as the grouping allows. On Eedi, probes are therefore always tested on constructs unseen in training; on ConceptKT, all options of a test question are unseen. Because constructs have unequal sizes, some rare classes are absent from individual test folds. We report the mean over folds and the standard deviation across folds.

\paragraph{Baselines.}
\emph{Chance} macro-$F_1$ is that of a predictor that guesses labels at random in proportion to the class frequencies: approximately $1/K$ for $K$ classes and $0.5$ for binary correctness. \emph{Majority-class accuracy} is the accuracy of always predicting the most frequent class: 9.6\% (Eedi misconception), 15.8\% (ConceptKT concept), 75\% (Eedi correctness), and 80\% (ConceptKT correctness). Accuracies should be read against the latter.

\paragraph{Emergence.}
For a task with per-layer macro-$F_1$ scores $s_0,\dots,s_L$, peak $s^\ast=\max_l s_l$, and chance level $c$, the emergence layer at threshold $r$ is
\begin{equation}
l_{\text{em}}(r) = \min\{\, l : s_l \ge c + r\,(s^\ast - c) \,\},
\end{equation}
reported as relative depth $l_{\text{em}}/L$; it is undefined when $s^\ast \le c$. We use $r{=}0.8$ in the main text and report $r \in \{0.6, 0.7, 0.9\}$ in Appendix~\ref{sec:app_emergence}.

\section{Dataset Construction}
\label{sec:app_data}

\subsection{Eedi}
\label{sec:app_eedi}
The Eedi release provides, for each of 1,869 questions, a construct (757 distinct), a subject, the question text, four options, the correct letter, and a misconception ID for each distractor where one has been annotated; the misconception catalogue contains 2,587 entries. Flattening gives 7,476 answer choices, 25\% of which are correct. Of the 5,607 distractors, 4,370 carry a misconception label, drawn from 1,604 distinct IDs whose frequency is highly skewed: most IDs occur only once or twice. Applying a minimum of 15 instances per class retains exactly 23 IDs and 565 instances from 391 questions, 188 constructs, and 69 subjects (Table~\ref{tab:app_classes_eedi}). The categories cover equation solving, rounding, order of operations, negative numbers, transformations, sequences, and fractions. The most frequent class (54 instances) gives the 9.6\% majority-class accuracy.

\begin{table}[!htbp]
\centering
\scriptsize
\setlength{\tabcolsep}{3pt}
\renewcommand{\arraystretch}{0.95}
\begin{tabular}{r p{0.70\linewidth} r}
\toprule
\textbf{ID} & \textbf{Misconception} & \textbf{$n$} \\
\midrule
1214 & When solving an equation, uses the same operation rather than the inverse. & 54 \\
1379 & Rounds down instead of up & 43 \\
2316 & Mixes up squaring and multiplying by 2 or doubling & 38 \\
1507 & Carries out operations from left to right regardless of priority order & 36 \\
1990 & Fails to reflect across mirror line & 33 \\
1880 & Mixes up greater than and less than symbols & 32 \\
1597 & Believes multiplying two negatives gives a negative answer & 27 \\
2392 & Rounds to the wrong degree of accuracy (rounds too much) & 27 \\
1248 & Rounds to the wrong degree of accuracy (rounds too little) & 22 \\
77 & Does not follow the arrows through a function machine, changes the order of the operations asked. & 22 \\
220 & Only multiplies the first term in the expansion of a bracket & 22 \\
1072 & Multiplies by the index & 20 \\
2481 & When multiplying decimals, divides by the wrong power of 10 when reinserting the decimal & 19 \\
2359 & Believes division is commutative & 19 \\
2355 & Does not know how to find order of rotational symmetry & 19 \\
974 & Believes multiplying a positive by a negative gives a positive answer & 18 \\
2271 & When solving a problem that requires an inverse operation (e.g. missing number problems), does the original operation & 18 \\
1988 & Rounds up instead of down & 17 \\
108 & Uses only the first two terms of a sequence to work out a term-to-term rule & 16 \\
31 & Does not understand bar modelling in algebra & 16 \\
217 & When adding fractions, adds the numerators and denominators & 16 \\
1198 & Does not recognise that a linear sequence must increase or decrease by same amount & 16 \\
113 & Does not recognise difference of two squares & 15 \\
\bottomrule
\end{tabular}
\caption{\textbf{The 23 Eedi misconceptions} with $\geq 15$ labelled distractors ($n$ = instances; 565 in total).}
\label{tab:app_classes_eedi}
\end{table}

\subsection{Worked Example}
\label{sec:app_example}
As an example, take the Eedi question \emph{``Solve $2n+3 \le 3n-7$''} from the construct \emph{Solve linear inequalities in one variable where the unknown appears on both sides}. Its four options are:
\begin{itemize}
    \setlength{\itemsep}{0pt}\setlength{\parskip}{0pt}\setlength{\parsep}{0pt}
    \item[\textbf{A.}] $2 \le n$ \quad misconception 1214, \emph{``When solving an equation, uses the same operation rather than the inverse''};
    \item[\textbf{B.}] $10 \le n$ \quad correct answer;
    \item[\textbf{C.}] $n \le 10$ \quad misconception 811, \emph{``Believes they can multiply/divide through by a negative number without changing the direction of the inequality''};
    \item[\textbf{D.}] $-4 \le n$ \quad misconception 1214 (as A).
\end{itemize}
\begin{itemize}
    \setlength{\itemsep}{0pt}\setlength{\parskip}{0pt}\setlength{\parsep}{0pt}
    \item \textbf{Misconception, Condition B.} Options A and D each yield one instance labelled 1214, whose prompt contains the construct, the stem, and that option; the probe reads the hidden state at the option's last token.
    \item \textbf{Misconception, Condition A.} The same two instances are represented by the single stem-only activation (construct and stem, no option), with label 1214, so the probe must predict the misconception without seeing either distractor.
    \item \textbf{Binary correctness.} All four options yield instances: B labelled correct, and A, C, and D labelled incorrect.
    \item \textbf{Correct-letter control.} The stem-only activation is labelled B.
\end{itemize}
Because folds are grouped by construct, no question from this construct appears in training when this question is tested.

\subsection{ConceptKT/MathQA}
\label{sec:app_ckt}
ConceptKT \cite{kang-etal-2026-conceptkt} provides, for MathQA problems \cite{amini-etal-2019-mathqa}, the list of concepts each problem requires (\texttt{associated\_concepts}) and, for a subset of student responses, the concepts the student was missing. ConceptKT question IDs index the concatenation of the MathQA train, dev, and test splits, which we use to retrieve the problem text, the five options, and the correct letter for 4,036 questions (20,180 choices, 20\% correct). Each question is labelled with the first entry of its \texttt{associated\_concepts} list, its primary concept; there are 52 distinct primary concepts, and we keep the 25 most frequent (Table~\ref{tab:app_classes_ckt}), covering 3,846 questions. Every distractor of a question receives the question's concept, giving 15,384 instances; folds are grouped by question, so all five options of a question are in the same fold. The most frequent concept (\emph{Ratio Calculation}, 607 questions) gives the 15.8\% majority-class accuracy. The prompt's topic line holds the MathQA category rather than the concept (Appendix~\ref{sec:app_prompts}); in about 10\% of instances a concept word also occurs naturally in the problem text (e.g.\ \emph{profit} in a \emph{Profit} problem).

\begin{table}[!htbp]
\centering
\scriptsize
\setlength{\tabcolsep}{3pt}
\renewcommand{\arraystretch}{0.95}
\begin{tabular}{p{0.58\linewidth} r r}
\toprule
\textbf{Concept} & \textbf{Questions} & \textbf{Instances} \\
\midrule
Ratio Calculation & 607 & 2,428 \\
Distance-Time-Speed & 544 & 2,176 \\
Expression and Operations with Symbols & 481 & 1,924 \\
Workload-Time-Speed & 338 & 1,352 \\
Arithmetic Mean & 330 & 1,320 \\
Basic Math Operations & 310 & 1,240 \\
Area Calculation & 194 & 776 \\
Profit & 183 & 732 \\
Simple Interest & 121 & 484 \\
Probability & 103 & 412 \\
Mixed Solution Concentration & 71 & 284 \\
Volume Calculation & 71 & 284 \\
Inclusion-Exclusion Principle & 64 & 256 \\
Arithmetic Sequence/Series & 58 & 232 \\
Permutations and Combinations & 47 & 188 \\
Discount Problem & 47 & 188 \\
Direct Proportion and Inverse Proportion & 40 & 160 \\
Unit Conversion & 36 & 144 \\
Perimeter Calculation & 35 & 140 \\
Prime Factorization & 33 & 132 \\
Plane Geometry & 29 & 116 \\
Function & 28 & 112 \\
Compound Interest & 27 & 108 \\
Graphs of Cartesian Coordinates and Linear Equations & 25 & 100 \\
Relative Speed & 24 & 96 \\
\bottomrule
\end{tabular}
\caption{\textbf{The 25 ConceptKT concepts} used as labels (primary associated concept; 3,846 questions, 15,384 distractor instances).}
\label{tab:app_classes_ckt}
\end{table}

\section{Additional Results}
\label{sec:app_results}

\begin{figure*}[t]
  \centering
  \includegraphics[width=0.92\linewidth]{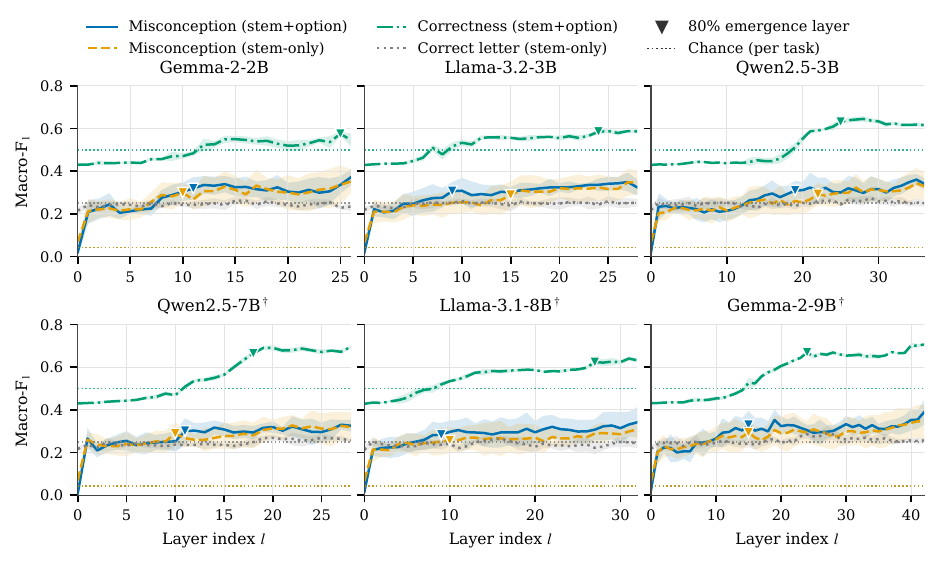}
  \caption{\textbf{Layer-by-layer linear-probe macro-$F_1$ on Eedi over absolute layer indices}, with $\pm1$ standard deviation across the 5 cross-validation folds (shaded). Dotted lines: chance per task; $\blacktriangledown$: 80\% emergence layer. $^\dagger$4-bit weights.}
  \label{fig:app_eedi_sd}
\end{figure*}

\subsection{Layer-wise Curves and Cross-Validation Variance}
\label{sec:app_layers}
Figure~\ref{fig:app_eedi_sd} shows the curves of Figure~\ref{fig:trajectories} over absolute layer indices, with $\pm1$ standard deviation across the five folds. Three patterns are visible in every model. First, misconception probes (blue and orange) rise steeply in the first few layers, then improve slowly and usually peak in the last few layers; the stem-only curve tracks the stem+option curve closely throughout. Second, binary correctness (green) stays close to its 0.5 chance line in early layers and rises in the middle-to-late layers. Third, the stem-only correct-letter control (grey) never moves far from its 0.25 chance line. Fold-to-fold standard deviation is up to 0.035 for the correctness tasks and up to 0.09 for the 23-way misconception tasks, reflecting the small per-class test counts (about three instances per class per fold). Peak differences of a few hundredths between models or conditions are therefore within fold variability, which is why we compare models and conditions qualitatively.

\subsection{ConceptKT/MathQA Results}
\label{sec:app_ckt_results}
\begin{figure*}[t]
  \centering
  \includegraphics[width=0.92\linewidth]{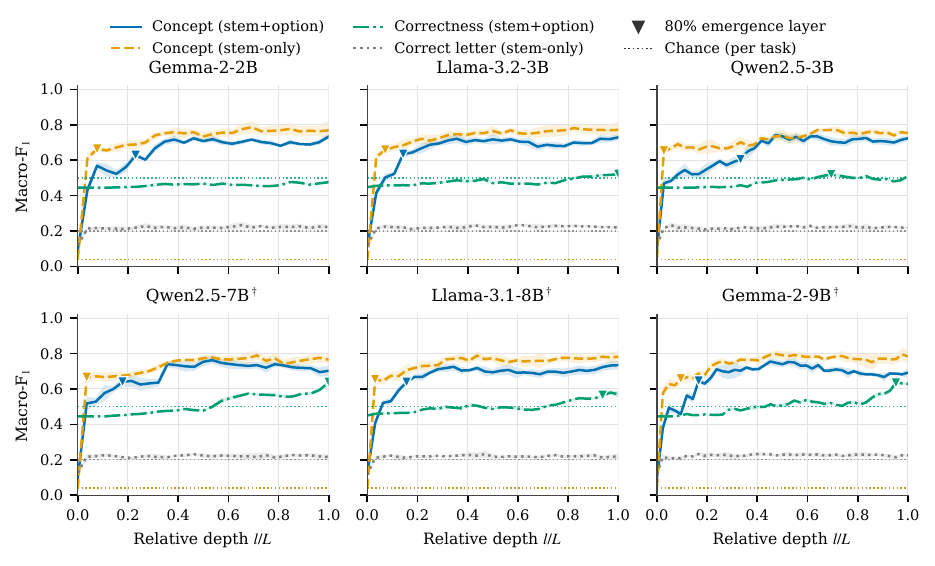}
  \caption{\textbf{Layer-wise linear-probe macro-$F_1$ on ConceptKT/MathQA across relative depth $l/L$}, with $\pm1$ standard deviation across folds (shaded). Concept = primary associated concept (25 classes). Dotted lines: chance per task; $\blacktriangledown$: 80\% emergence layer (omitted when the peak does not exceed chance). $^\dagger$4-bit weights.}
  \label{fig:app_ckt_traj}
\end{figure*}

Table~\ref{tab:app_detail_ckt} gives the complete ConceptKT results and Figure~\ref{fig:app_ckt_traj} the layer-wise curves. Concept probes rise steeply within the first few layers and keep improving into the last layers; the stem-only curve lies above the stem+option curve at most depths. Correctness probes stay near the 0.5 chance line for most of the network and rise only in the final layers, and their accuracy (80.1\%--82.0\%) stays close to the 80\% majority rate; for \model{Gemma-2-2B} correctness never exceeds chance, so its emergence layer is undefined. The 5-way stem-only correct-letter control peaks at 0.231--0.236 (chance 0.20). Fold-to-fold standard deviation is at most 0.053 for the concept tasks and 0.021 for the correctness tasks, smaller than on Eedi because ConceptKT has about 27 times as many label-task instances.

\begin{table*}[t]
\centering
\small
\begin{adjustbox}{max width=\textwidth}
\begin{tabular}{llrccccc}
\toprule
\textbf{Model} & \textbf{Task (condition)} & \textbf{$N$} & \textbf{Chance $F_1$} & \textbf{Peak layer} & \textbf{Peak $F_1$} & \textbf{Peak acc.\ (\%)} & \textbf{Emergence (80\%)} \\
\midrule
\multirow{4}{*}{\model{Gemma-2-2B}} & Concept (B) & 15,384 & 0.040 & 26 & 0.7338 & 81.57 & 6 (0.23) \\
 & Concept (A) & 15,384 & 0.040 & 18 & 0.7865 & 85.08 & 2 (0.08) \\
 & Correctness (B) & 20,180 & 0.500 & 22 & 0.4768 & 80.07 & n/a \\
 & Correct letter (A, control) & 4,036 & 0.200 & 17 & 0.2333 & 23.49 & 7 (0.27) \\
\midrule
\multirow{4}{*}{\model{Llama-3.2-3B}} & Concept (B) & 15,384 & 0.040 & 28 & 0.7290 & 81.69 & 4 (0.14) \\
 & Concept (A) & 15,384 & 0.040 & 23 & 0.7812 & 85.00 & 2 (0.07) \\
 & Correctness (B) & 20,180 & 0.500 & 28 & 0.5213 & 80.36 & 28 (1.00) \\
 & Correct letter (A, control) & 4,036 & 0.200 & 17 & 0.2355 & 23.84 & 12 (0.43) \\
\midrule
\multirow{4}{*}{\model{Qwen2.5-3B}} & Concept (B) & 15,384 & 0.040 & 17 & 0.7408 & 82.51 & 12 (0.33) \\
 & Concept (A) & 15,384 & 0.040 & 25 & 0.7713 & 85.05 & 1 (0.03) \\
 & Correctness (B) & 20,180 & 0.500 & 25 & 0.5195 & 80.17 & 25 (0.69) \\
 & Correct letter (A, control) & 4,036 & 0.200 & 2 & 0.2336 & 23.49 & 2 (0.06) \\
\midrule
\multirow{4}{*}{\model{Qwen2.5-7B}$^\dagger$} & Concept (B) & 15,384 & 0.040 & 15 & 0.7625 & 84.25 & 5 (0.18) \\
 & Concept (A) & 15,384 & 0.040 & 20 & 0.7894 & 86.45 & 1 (0.04) \\
 & Correctness (B) & 20,180 & 0.500 & 28 & 0.6377 & 81.43 & 28 (1.00) \\
 & Correct letter (A, control) & 4,036 & 0.200 & 13 & 0.2310 & 23.41 & 3 (0.11) \\
\midrule
\multirow{4}{*}{\model{Llama-3.1-8B}$^\dagger$} & Concept (B) & 15,384 & 0.040 & 32 & 0.7342 & 82.72 & 5 (0.16) \\
 & Concept (A) & 15,384 & 0.040 & 14 & 0.7883 & 85.88 & 1 (0.03) \\
 & Correctness (B) & 20,180 & 0.500 & 31 & 0.5797 & 80.78 & 30 (0.94) \\
 & Correct letter (A, control) & 4,036 & 0.200 & 15 & 0.2322 & 23.39 & 13 (0.41) \\
\midrule
\multirow{4}{*}{\model{Gemma-2-9B}$^\dagger$} & Concept (B) & 15,384 & 0.040 & 19 & 0.7549 & 83.55 & 7 (0.17) \\
 & Concept (A) & 15,384 & 0.040 & 20 & 0.7971 & 86.51 & 4 (0.10) \\
 & Correctness (B) & 20,180 & 0.500 & 40 & 0.6343 & 82.01 & 40 (0.95) \\
 & Correct letter (A, control) & 4,036 & 0.200 & 7 & 0.2324 & 23.59 & 7 (0.17) \\
\bottomrule
\end{tabular}
\end{adjustbox}
\caption{\textbf{Detailed linear-probe results on ConceptKT/MathQA.} Emergence is given as layer index ($l/L$). Majority-class accuracy is 15.8\% for the concept task and 80\% for binary correctness; the correct-letter control has 5 classes. $^\dagger$4-bit weights.}
\label{tab:app_detail_ckt}
\end{table*}

\subsection{Emergence under Alternative Thresholds}
\label{sec:app_emergence}
The emergence layer depends on the threshold $r$. Table~\ref{tab:app_thresholds} reports the relative emergence depth at $r \in \{0.6, 0.7, 0.8, 0.9\}$ for Condition-B label and correctness probes on both datasets. Misconception emerges earlier than correctness in all six models for $r \le 0.8$, and the margin is largest at $r{=}0.6$ (0.04--0.21 vs.\ 0.43--0.61). At $r{=}0.9$ the ordering holds in four models and reverses in \model{Qwen2.5-3B} (0.92 vs.\ 0.69) and \model{Gemma-2-9B} (1.00 vs.\ 0.95), whose misconception curves keep improving until the final layers, so reaching 90\% of the peak happens late. On ConceptKT, the concept emerges before correctness at every threshold in the five models whose correctness probe exceeds chance. The ordering is thus robust for moderate thresholds and weakens on Eedi only when emergence is defined very close to the peak.

\begin{table}[!htbp]
\centering
\footnotesize
\setlength{\tabcolsep}{2.4pt}
\begin{adjustbox}{max width=\columnwidth}
\begin{tabular}{l cccc cccc}
\toprule
& \multicolumn{4}{c}{\textbf{Label (B)}} & \multicolumn{4}{c}{\textbf{Correctness (B)}} \\
\cmidrule(lr){2-5}\cmidrule(lr){6-9}
\textbf{Model} & 60 & 70 & 80 & 90 & 60 & 70 & 80 & 90 \\
\midrule
\multicolumn{9}{l}{\textit{Eedi (23 misconceptions)}} \\
\model{Gemma-2-2B} & 0.12 & 0.31 & 0.42 & 0.54 & 0.54 & 0.96 & 0.96 & 0.96 \\
\model{Llama-3.2-3B} & 0.14 & 0.21 & 0.32 & 0.64 & 0.43 & 0.75 & 0.86 & 0.86 \\
\model{Qwen2.5-3B} & 0.06 & 0.39 & 0.53 & 0.92 & 0.58 & 0.67 & 0.69 & 0.69 \\
\model{Qwen2.5-7B}$^\dagger$ & 0.04 & 0.04 & 0.39 & 0.43 & 0.61 & 0.64 & 0.64 & 0.68 \\
\model{Llama-3.1-8B}$^\dagger$ & 0.09 & 0.22 & 0.28 & 0.69 & 0.56 & 0.84 & 0.84 & 0.97 \\
\model{Gemma-2-9B}$^\dagger$ & 0.21 & 0.26 & 0.36 & 1.00 & 0.52 & 0.57 & 0.57 & 0.95 \\
\midrule
\multicolumn{9}{l}{\textit{ConceptKT/MathQA (25 concepts)}} \\
\model{Gemma-2-2B} & 0.08 & 0.08 & 0.23 & 0.31 & n/a & n/a & n/a & n/a \\
\model{Llama-3.2-3B} & 0.07 & 0.11 & 0.14 & 0.21 & 0.93 & 0.93 & 1.00 & 1.00 \\
\model{Qwen2.5-3B} & 0.03 & 0.11 & 0.33 & 0.42 & 0.67 & 0.69 & 0.69 & 0.69 \\
\model{Qwen2.5-7B}$^\dagger$ & 0.04 & 0.11 & 0.18 & 0.36 & 0.96 & 1.00 & 1.00 & 1.00 \\
\model{Llama-3.1-8B}$^\dagger$ & 0.06 & 0.09 & 0.16 & 0.19 & 0.94 & 0.94 & 0.94 & 0.97 \\
\model{Gemma-2-9B}$^\dagger$ & 0.05 & 0.12 & 0.17 & 0.24 & 0.93 & 0.95 & 0.95 & 0.95 \\
\bottomrule
\end{tabular}
\end{adjustbox}
\caption{\textbf{Emergence depth under alternative thresholds} (linear probes, Condition B). Relative depth $l/L$ of the first layer reaching $r\in\{60,70,80,90\}\%$ of the peak macro-$F_1$ above chance; the label is the misconception (Eedi) or concept (ConceptKT). n/a: peak not above chance. $^\dagger$4-bit weights.}
\label{tab:app_thresholds}
\end{table}

\subsection{Linear vs.\ MLP Probes}
\label{sec:app_parity}
\begin{figure}[!htbp]
  \centering
  \includegraphics[width=\columnwidth]{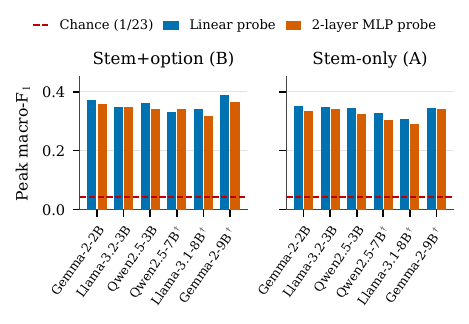}
  \caption{\textbf{Linear vs.\ 2-layer MLP probes on Eedi misconceptions} (peak macro-$F_1$). Dashed: uniform chance. $^\dagger$4-bit weights.}
  \label{fig:app_parity}
\end{figure}

Table~\ref{tab:app_parity} compares linear and 2-layer MLP probes on the label task for both datasets, and Figure~\ref{fig:app_parity} plots the Eedi comparison. The MLP never exceeds the linear probe by more than 0.012 macro-$F_1$ and is lower in 11 of 12 model--condition pairs, by up to 0.024; peak accuracy differs by $-3.8$ to $+0.7$ points (MLP minus linear). Extra non-linear capacity therefore yields no gain, consistent with misconception information being linearly accessible. This does not rule out non-linear structure that a larger probe or more training data could exploit; with 565 instances, a higher-capacity probe may also be more prone to overfitting. On ConceptKT, with 15,384 instances, the MLP is higher in all 12 model--condition pairs, by 0.019--0.032 under Condition B and 0.007--0.029 under Condition A; the gain is small relative to the gap above chance, so concept information is also largely linearly accessible.

\begin{table}[!htbp]
\centering
\footnotesize
\setlength{\tabcolsep}{3pt}
\begin{adjustbox}{max width=\columnwidth}
\begin{tabular}{l ccc ccc}
\toprule
& \multicolumn{3}{c}{\textbf{Stem+option (B)}} & \multicolumn{3}{c}{\textbf{Stem-only (A)}} \\
\cmidrule(lr){2-4}\cmidrule(lr){5-7}
\textbf{Model} & Linear & MLP & $\Delta$ & Linear & MLP & $\Delta$ \\
\midrule
\multicolumn{7}{l}{\textit{Eedi (23 misconceptions)}} \\
\model{Gemma-2-2B} & 0.371 & 0.359 & -0.012 & 0.351 & 0.334 & -0.018 \\
\model{Llama-3.2-3B} & 0.349 & 0.347 & -0.002 & 0.348 & 0.341 & -0.007 \\
\model{Qwen2.5-3B} & 0.360 & 0.340 & -0.020 & 0.344 & 0.325 & -0.019 \\
\model{Qwen2.5-7B}$^\dagger$ & 0.330 & 0.342 & +0.012 & 0.327 & 0.305 & -0.022 \\
\model{Llama-3.1-8B}$^\dagger$ & 0.342 & 0.318 & -0.024 & 0.308 & 0.292 & -0.016 \\
\model{Gemma-2-9B}$^\dagger$ & 0.389 & 0.366 & -0.023 & 0.346 & 0.341 & -0.005 \\
\midrule
\multicolumn{7}{l}{\textit{ConceptKT/MathQA (25 concepts)}} \\
\model{Gemma-2-2B} & 0.734 & 0.761 & +0.028 & 0.786 & 0.807 & +0.021 \\
\model{Llama-3.2-3B} & 0.729 & 0.761 & +0.032 & 0.781 & 0.798 & +0.016 \\
\model{Qwen2.5-3B} & 0.741 & 0.766 & +0.025 & 0.771 & 0.800 & +0.029 \\
\model{Qwen2.5-7B}$^\dagger$ & 0.763 & 0.784 & +0.022 & 0.789 & 0.807 & +0.017 \\
\model{Llama-3.1-8B}$^\dagger$ & 0.734 & 0.761 & +0.026 & 0.788 & 0.805 & +0.017 \\
\model{Gemma-2-9B}$^\dagger$ & 0.755 & 0.774 & +0.019 & 0.797 & 0.804 & +0.007 \\
\bottomrule
\end{tabular}
\end{adjustbox}
\caption{\textbf{Linear vs.\ 2-layer MLP probes}: peak macro-$F_1$ for the label task and difference $\Delta$ = MLP $-$ linear. $^\dagger$4-bit weights.}
\label{tab:app_parity}
\end{table}

\subsection{Relation to Model Size}
\label{sec:app_scale}
\begin{figure}[!htbp]
  \centering
  \includegraphics[width=\columnwidth]{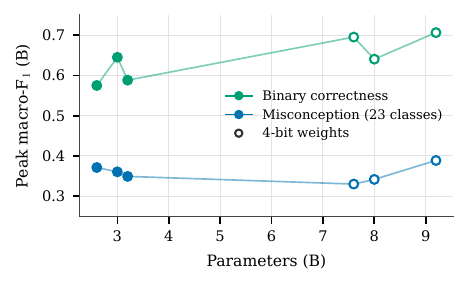}
  \caption{\textbf{Peak Condition-B macro-$F_1$ vs.\ parameter count} on Eedi for the misconception task and binary correctness. Hollow markers: 4-bit weights.}
  \label{fig:app_scale}
\end{figure}

Figure~\ref{fig:app_scale} plots peak Condition-B macro-$F_1$ against parameter count. Misconception $F_1$ is 0.371, 0.361, 0.349, 0.330, 0.342, and 0.389 for the 2.6B, 3.0B, 3.2B, 7.6B, 8.0B, and 9.2B models: the largest model is best, but the 7--8B models score below the 2--3B models. Correctness $F_1$ shows a clearer upward trend, from 0.575 (2.6B) to 0.706 (9.2B), with dips at 3.2B and 8.0B. We therefore see no consistent scaling of misconception decodability over this range. Since the three largest models are also the 4-bit ones (hollow markers), model size and weight precision cannot be separated here.

\section{Use of AI Assistants}
\label{sec:app_ai}
AI assistants (ChatGPT, Gemini, Claude) were used for language editing, restructuring text, and limited programming assistance (debugging and code refinement). All code, experiments, analyses, interpretations, and the final manuscript were reviewed and verified by the authors, who take full responsibility for the content of this work.
\end{document}